\RequirePackage{fix-cm}
\documentclass[journal,twoside,web]{ieeecolor}
\usepackage{generic}

\usepackage{cite}
\usepackage{amsmath,amssymb,amsfonts}
\usepackage{graphicx}
\usepackage{booktabs}
\usepackage{multirow}
\usepackage{array}
\usepackage{url}
\usepackage{textcomp}
\usepackage{placeins}
\usepackage{etoolbox}
\AtBeginDocument{\apptocmd{\thebibliography}{\fontsize{7.5}{8.5}\selectfont}{}{}}
\usepackage{balance}
\usepackage{hyperref}
\hypersetup{hidelinks,pdftitle={SFL-Net: Dependence-Regularized Latent Pathways for MRI-to-tau-PET Synthesis},pdfauthor={Agamdeep S. Chopra et al.}}
\usepackage{microtype}
\DeclareRobustCommand{\msemark}{\ensuremath{^{\dagger}}}
\DeclareRobustCommand{\vccmark}{\ensuremath{^{\ddagger}}}
\DeclareRobustCommand{\agmark}{\ensuremath{^{\ast}}}
\DeclareRobustCommand{\spademark}{\ensuremath{^{\circledast}}}
\DeclareRobustCommand{\sfmark}{\ensuremath{^{\odot}}}
\DeclareRobustCommand{\rmark}{\ensuremath{^{\curlyvee}}}

\def\BibTeX{{\rm B\kern-.05em{\sc i\kern-.025em b}\kern-.08em T\kern-.1667em\lower.7ex\hbox{E}\kern-.125emX}}
\providecommand{\refname}{References}
\title{SFL-Net: Dependence-Regularized Latent Pathways for Multi-Contrast MRI-to-tau-PET Synthesis and Component Attribution}

\author{\centering
Agamdeep~S.~Chopra, Caitlin~Neher, Tianyi~Ren, Juampablo~E.~Heras~Rivera,\\
Hesamoddin~Jahanian, and Mehmet~Kurt\endgraf
\thanks{\raggedright This work was supported in part by the U.S. Department of Energy, Office of Science, Office of Advanced Scientific Computing Research, DOE Computational Science Graduate Fellowship under Award No. DE-SC0024386 to Juampablo E. Heras Rivera.}
\thanks{\raggedright Agamdeep S. Chopra, Caitlin Neher, Tianyi Ren, Juampablo E. Heras Rivera, and Mehmet Kurt are with the Department of Mechanical Engineering, University of Washington, Seattle, WA, USA (e-mail: achopra4@uw.edu; neherc@uw.edu; tr1@uw.edu; jehr@uw.edu; mkurt@uw.edu).}
\thanks{\raggedright Hesamoddin Jahanian is with the Department of Radiology, University of Washington, Seattle, WA, USA (e-mail: hesamj@uw.edu).}
\thanks{\raggedright Caitlin Neher and Tianyi Ren contributed equally. Correspondence should be addressed to Agamdeep S. Chopra (e-mail: achopra4@uw.edu).}}

\begin{document}
\maketitle
\begin{abstract}
\raggedright
Structural MRI to tau-PET synthesis is weakly constrained because T1-weighted and FLAIR MRI provide anatomical and disease-associated structural information but do not directly measure tau tracer binding. We investigate whether an explicitly interrogable latent bottleneck can be imposed on this problem while retaining competitive synthesis performance. SFL-Net partitions a vector-quantized representation into shared, T1-specific, FLAIR-specific, and cross-source complementary pathways and uses latent-derived structural conditioning rather than direct full-resolution encoder--decoder feature skips. We evaluated SFL-Net and capacity-matched VAE, VQ-VAE, U-Net, and SPADE-U-Net variants using 605 training and 83 held-out internal validation subjects pooled from ADNI-3 and OASIS-3. SFL-Net achieved the lowest raw PET MAE and MSE and the highest raw PET PSNR among the implemented models, whereas SPADE-U-Net variants were stronger on several SUVR similarity, high-uptake, and threshold-based staging metrics. Exact coalitional Shapley attribution assigned the largest mean marginal contribution to the cross-source pathway across the evaluated reconstruction and staging utilities, and the subject-level attribution outputs satisfied the Shapley efficiency identity to floating-point precision. This attribution is operational and conditional on a retained structure-only pathway; it does not establish identifiable information-theoretic synergy or the incremental value of two MRI contrasts relative to matched single-contrast models. All evaluated model families showed dynamic-range compression at higher tau burden, and Braak-derived staging performance was close to a constant Stage~1 reference on several metrics. These findings support SFL-Net as a framework for component-level auditing of multi-input synthesis under internal validation rather than as a clinically validated replacement for acquired tau-PET.
\end{abstract}

\section{Introduction}
\IEEEPARstart{T}{au} positron emission tomography (tau-PET) provides in-vivo information about neurofibrillary tau pathology and is closely related to Alzheimer's disease (AD) stage and clinical decline \cite{jack2018niaaa,burnham2024flortaucipir,chen2021taupet}. Its routine use is constrained by radiotracer production and administration, ionizing radiation, acquisition time, cost, and specialized PET infrastructure. Structural MRI is more widely available and motivates investigation of MRI to PET synthesis, but availability alone does not imply that MRI contains a subject-specific molecular measurement. T1-weighted MRI primarily depicts anatomy and atrophy, while FLAIR emphasizes white-matter hyperintensities, vascular injury, and other tissue-signal abnormalities. Neither directly measures tau binding. Accordingly, structural MRI to tau-PET synthesis should be treated as estimation of a conditional imaging pattern from indirect structural correlates, not recovery of otherwise observed molecular information. There is nevertheless an empirical basis for a statistical mapping: tau-PET burden is associated with cortical thinning and longitudinal atrophy, although these relationships are incomplete and heterogeneous across disease stage and region \cite{gordon2018tauatrophy,lee2024synthetic}.

Existing tau-PET synthesis studies span input settings with very different information content. In a large cross-modal study, Lee et al. synthesized tau-PET from T1-weighted MRI, FDG-PET, and amyloid-PET and found the T1-based mapping substantially less accurate than PET-conditioned mappings, underscoring the weaker observability of molecular tau signal from structure alone \cite{lee2024synthetic}. Recent structural MRI work has also moved beyond simple convolutional regression: Moon et al. evaluated T1-weighted MRI to tau-PET synthesis across U-Net, UNETR, SwinUNETR, CycleGAN, and Pix2Pix families \cite{moon2026cyclic}, while pathology-aware diffusion models have been developed for MRI to PET translation more broadly \cite{li2024pasta}. Multi-input MRI synthesis also has prior examples of source-specific and common latent representations, including Hi-Net and feature-disentanglement approaches that separate shared and source-specific information \cite{zhou2020hinet,fei2021disentanglement}. Thus, shared/specific factorization by itself is not the novelty claimed here.

The research question is how a trained two-input predictor depends on prescribed latent routes under a declared replacement intervention. Here, \emph{source factorization} is an operational name for the architecture and dependence regularization, not a recovered decomposition of source information. We use shared-designated, T1-only-input, FLAIR-only-input, and joint-input to describe the latent pathways. These labels specify input access and intended training roles; they do not establish exclusivity, orthogonality, or biological meaning. A joint-input pathway may use predominantly one contrast or duplicate information carried elsewhere.

Direct encoder--decoder skips preserve spatial detail but complicate an analysis restricted to bottleneck components because information also reaches the output through unexamined routes. Skips are not intrinsically unsuitable for synthesis or attribution: they can themselves be included as intervention units. SFL-Net restricts direct feature bypass to define a particular latent intervention experiment, retaining a latent-derived structural pathway. For example, branch replacement can reveal whether a fitted decoder relies on its joint-input route to reduce reconstruction error under fixed structural context. This is a research inspection question, not a basis for choosing a patient's MRI acquisitions or localizing tau pathology.

Synthetic PET changes the representation of the observed MRI; it adds no independently measured patient information. Early molecular change without structural change and non-AD structural injury are plausible failure settings for a learned MRI--tau association. They are not subgroup findings established here. We therefore use image reconstruction, regional summaries, and adapted threshold-derived staging to characterize limitations rather than to establish clinical substitution.

This work brings together vector-quantized latent pathways, dependence regularization, latent-derived structural conditioning, and exact coalition attribution in a common synthesis framework. The prescribed pathways provide explicit units for model interrogation, while the evaluation connects reconstruction quality with regional uptake and ordinal imaging summaries. The contribution lies in this integrated design and the resulting characterization of decoder dependence across subjects. Comparative findings refer to the configurations implemented under the study protocol.

\begin{figure*}[t]
\centering
\includegraphics[width=0.96\textwidth,height=0.69\textheight,keepaspectratio]{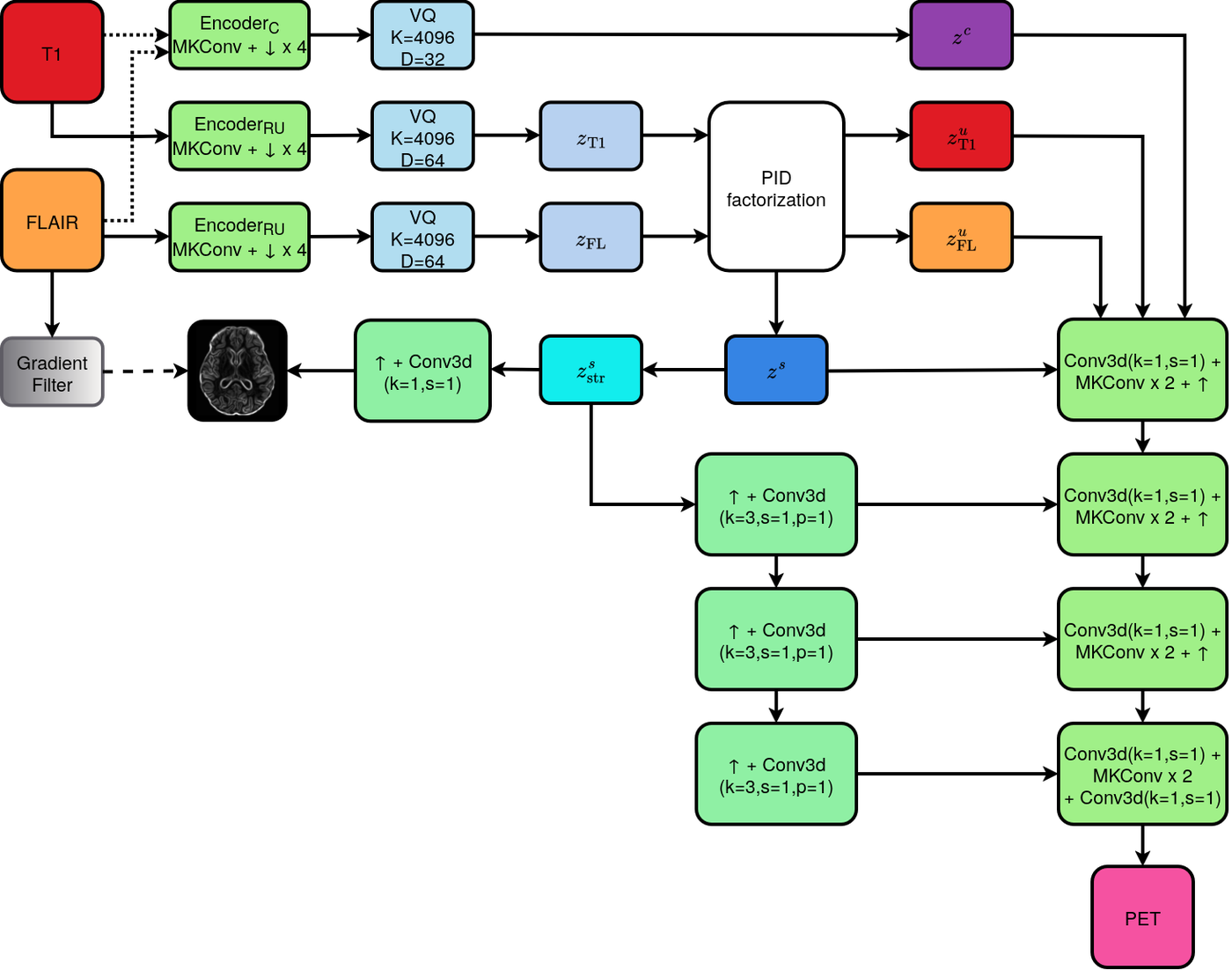}
\caption{SFL-Net architecture for structural MRI-to-tau-PET synthesis. The two Encoder$_{\rm RU}$ applications share weights; Encoder$_{\rm C}$ processes both T1-weighted and FLAIR inputs. $K$ and $D$ denote codebook size and code-vector dimension. Quantized representations are organized into shared-designated, T1-only-input, FLAIR-only-input, and joint-input pathways. The group labeled ``PID factorization'' denotes prescribed latent partitioning and dependence regularization, rather than a formal PID estimator or an additional learned forward transform. The diagram depicts FLAIR-derived edge supervision during training; PET decoding receives latent-derived structural maps. $\uparrow$ denotes upsampling and $\downarrow$ denotes downsampling. The joint-input label ``complementary'' denotes an architectural role rather than identifiable information-theoretic synergy.}
\label{fig:sfl}
\end{figure*}

% AUTHOR CHECKS: see AUTHOR_CHECKLIST.md; resolve configuration and evaluation details before submission.
\section{Method}
\subsection{SFL-Net Architecture}
SFL-Net contains two weight-shared single-source encoder applications, a separate cross-source encoder, vector-quantization layers, and a structure-conditioned latent decoder (Fig.~\ref{fig:sfl}). The T1-weighted and FLAIR inputs are processed independently by the shared/source-specific encoder, while the cross-source encoder receives their concatenation. The resulting latent representation is organized as
\begin{equation}
 z_\mathrm{T1}\rightarrow[z^s_\mathrm{T1},z^u_\mathrm{T1}],\quad
 z_\mathrm{FL}\rightarrow[z^s_\mathrm{FL},z^u_\mathrm{FL}],\quad
 z_\mathrm{C}\rightarrow z^c.
\end{equation}
The labels denote intended architectural roles, not measured source-information atoms. The shared factors $z^s_\mathrm{T1}$ and $z^s_\mathrm{FL}$ are encouraged to agree across sources; $z^u_\mathrm{T1}$ and $z^u_\mathrm{FL}$ are source-specific residual pathways whose cross-dependence is discouraged; and $z^c$ is produced from the joint T1/FLAIR input and regularized to remain distinct from the single-source partitions. We refer to $z^c$ as the \emph{joint-input} pathway; the label ``complementary'' refers to this same branch and its intended architectural role. The architecture does not establish that $z^c$ contains formal target-relative synergy or information unavailable from either source in the information-theoretic sense.

The single-source vector-quantization pathways use 4096-entry codebooks with 64-dimensional code vectors, and the joint cross-source pathway uses a 4096-entry codebook with 32-dimensional code vectors. Vector-quantized models were trained with 20 warmup epochs in which quantization was disabled, followed by $k$-means codebook initialization and exponential-moving-average codebook updates \cite{oord2017vqvae}.

\begin{figure}[t]
\centering
\includegraphics[width=0.94\columnwidth]{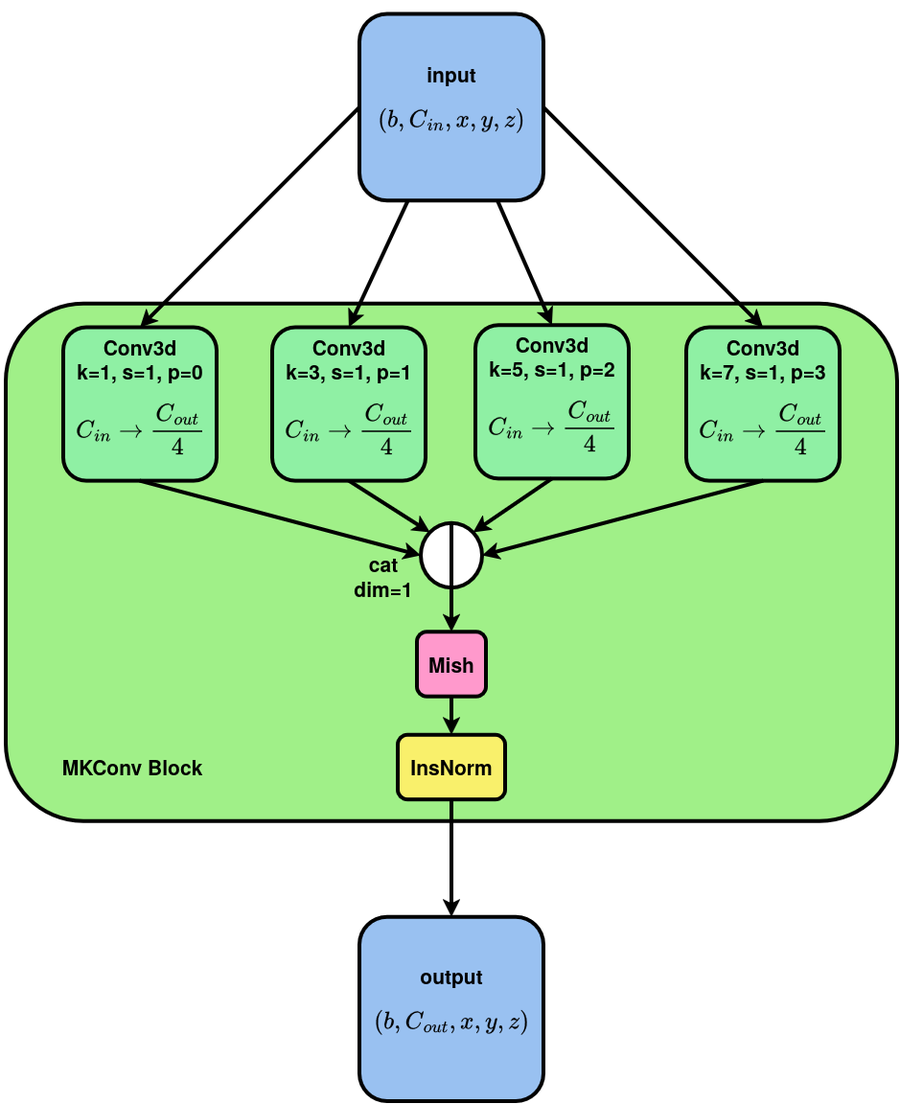}
\caption{Multi-kernel convolution (MKConv) block used in the convolutional backbones. Parallel 3D convolutions with kernel sizes 1, 3, 5, and 7 each produce one quarter of the requested output channels; the branch outputs are concatenated, followed by Mish activation and instance normalization.}
\label{fig:mkconv}
\end{figure}

\subsection{Latent Source Factorization}
The source-factorization objective shapes the dependence pattern among the predefined latent pathways. It is conceptually motivated by the distinction between shared, source-specific, and interaction-dependent information, but it is not a formal PID estimator \cite{williams2010pid,GriffithHo2015}.

For pre-quantized latent $z_P$ and codebook $\mathcal{E}_P=\{e_{P,k}\}_{k=1}^{K_P}$, a differentiable soft assignment is
\begin{equation}
\pi_P(k\mid z_P)=\mathrm{softmax}_k\!\left(-\frac{\|z_P-e_{P,k}\|_2^2}{\tau}\right),
\end{equation}
where $\tau$ controls assignment sharpness \cite{jang2017categorical,maddison2017concrete}. For $M$ paired latent samples from partitions $A$ and $B$, we form a soft contingency distribution by averaging the outer products of the two assignment vectors,
\begin{equation}
\widehat p_{AB}(k,\ell)=\frac{1}{M}\sum_{n=1}^{M}\pi_A(k\mid z_A^{(n)})\pi_B(\ell\mid z_B^{(n)}).
\end{equation}
Each outer product sums to one; therefore $\widehat p_{AB}$ is normalized, and its marginals are obtained by summation,
\begin{equation}
\widehat p_A(k)=\sum_{\ell=1}^{K_B}\widehat p_{AB}(k,\ell),\qquad
\widehat p_B(\ell)=\sum_{k=1}^{K_A}\widehat p_{AB}(k,\ell).
\end{equation}
This construction is a differentiable soft co-assignment surrogate. If the assignment vectors are interpreted probabilistically, the outer product corresponds to independent soft code draws conditional on each paired latent observation; it does not assume marginal independence between partitions $A$ and $B$, nor should it be interpreted as an exact probabilistic model of the original continuous latents.

For the corresponding normalized discrete distributions, mutual information is
\begin{equation}
I(A;B)=\sum_{k,\ell:\widehat p_{AB}(k,\ell)>0}\widehat p_{AB}(k,\ell)
\log\frac{\widehat p_{AB}(k,\ell)}{\widehat p_A(k)\widehat p_B(\ell)}.
\end{equation}
The finite-alphabet bound $I(A;B)\leq \log(\min(K_A,K_B))$ motivates the capacity-normalized dependence regularizer
\begin{equation}
\mathrm{cNMI}(A;B)=\frac{I(A;B)}{\log(\min(K_A,K_B))}.
\end{equation}
For $K_A,K_B>1$, the mathematical estimator above satisfies $\mathrm{cNMI}\in[0,1]$ \cite{cover2006elements}. The training implementation used a small numerical $\epsilon$ when evaluating logarithms (and the denominator) to avoid undefined floating-point operations. The strict bound refers to the unstabilized normalized discrete estimator; the implemented cNMI should be interpreted as a numerically stabilized training regularizer rather than as a calibrated information measure. Unlike entropy-normalized NMI, the denominator is fixed by codebook capacity rather than realized minibatch entropy \cite{strehl2002cluster,kvalseth2017normalized}.

The source-factorization loss is
\begin{align}
\mathcal{L}_{\mathrm{SF}}={}&
\lambda_u\,\mathrm{cNMI}(z^u_\mathrm{T1};z^u_\mathrm{FL})
+\lambda_s\mathcal{R}_s \nonumber\\
&+\lambda_{us}\sum_{i,j}\mathrm{cNMI}(z^u_i;z^s_j)
+\lambda_{cs}\sum_j\mathrm{cNMI}(z^c;z^s_j) \nonumber\\
&+\lambda_{cu}\sum_i\mathrm{cNMI}(z^c;z^u_i),\\
\mathcal{R}_s={}&1-\mathrm{cNMI}(z^s_\mathrm{T1};z^s_\mathrm{FL}) \nonumber\\
&+\lambda_{\mathrm{align}}\|z^s_\mathrm{T1}-z^s_\mathrm{FL}\|_2^2.
\end{align}
where $i,j\in\{\mathrm{T1},\mathrm{FL}\}$ and $\lambda_{\mathrm{align}}=1$. These terms encourage agreement of the shared factors and reduce pairwise dependence among factors assigned different roles. They neither estimate PID atoms nor establish source-exclusive or target-relevant information. Pairwise penalties do not exclude higher-order dependence or high-variance nuisance representations. The estimator scale depends on codebook capacity and the observation pool, not only on the underlying sources. Gradient accumulation increases effective optimization batch size but does not by itself enlarge the observation pool in an individual MI estimate.

The anti-collapse regularizer for the non-shared pathways takes the form
\begin{equation}
\mathcal{L}_{\mathrm{vf}}(z)=\frac{1}{C}\sum_{j=1}^{C}\max\!\left(0,\sigma_0-\mathrm{Std}(z_j)\right),
\end{equation}
and define
\begin{equation}
\mathcal{L}_{\mathrm{vf,tot}}=\mathcal{L}_{\mathrm{vf}}(z^u_\mathrm{T1})+\mathcal{L}_{\mathrm{vf}}(z^u_\mathrm{FL})+\mathcal{L}_{\mathrm{vf}}(z^c).
\end{equation}
This term is intended to discourage low variance; it does not establish target relevance or prove that collapse was avoided. Random input swapping and stochastic shared-factor mixing were additionally used during training,
\begin{equation}
z^s=\rho z^s_\mathrm{T1}+(1-\rho)z^s_\mathrm{FL},\qquad \rho\sim\mathcal{U}(0,1),
\end{equation}
Stochastic mixing is restricted to training. A convex combination after quantization need not be a codebook entry, whereas mixing before requantization defines a different operation. These alternatives should not be treated as equivalent.

\subsection{Structure-Conditioned Latent Decoding}
Direct feature skips introduce routes outside a bottleneck-only intervention \cite{ronneberger2015unet}. SFL-Net instead uses structural maps predicted from a fixed channel slice of the shared representation. Edge supervision constrains this pathway but does not guarantee that it transmits anatomy alone. This slice is supervised against an input-derived structural edge map during training.
We thus reserve a fixed set of channels $\mathcal{I}_{\mathrm{str}}$ within $z^s$ for structural conditioning and define
\begin{equation}
z_{\mathrm{str}}
=
z^s[:,\mathcal{I}_{\mathrm{str}},\cdot],
\end{equation}
where the reserved channel subset is fixed across training and inference. The decoder is initialized at the coarsest scale by
\begin{equation}
d_L=\phi_L\!\left(\eta(z^s,z^u_\mathrm{T1},z^u_\mathrm{FL},z^c)\right),
\end{equation}
where $\eta(\cdot)$ denotes latent fusion. At decoder level $\ell$, a scale-matched structural conditioning map is predicted from $z_{\mathrm{str}}$ and fused with the upsampled decoder feature:
\begin{equation}
\tilde{q}_{\ell}
=
\psi_{\ell}(z_{\mathrm{str}}),
\qquad
d_{\ell}
=
\phi_{\ell}\!\left(
\mathrm{Up}(d_{\ell+1})\Vert \tilde{q}_{\ell}
\right),
\end{equation}
where $\Vert$ indicates concatenation along the channel dimension, $\phi_{\ell}$ denotes the decoder block, and $\psi_{\ell}$ corresponds to the complete transform producing the structural map at that decoder level, including preceding scale transforms if implemented as a cascade. This notation does not imply independent projection heads.

For structural supervision, a fixed 3D Sobel edge magnitude map from an MRI input is min--max normalized to obtain $\hat{M}$ \cite{sobel1968,brats2022-transmorph}. Structural consistency is enforced in $z_{\mathrm{str}}$ by
\begin{equation}
\mathcal{L}_{\mathrm{edge}}
=
\left\|
\tilde{M}-\hat{M}
\right\|_2^2,
\qquad
\tilde{M} = \theta(z_{\mathrm{str}}).
\end{equation}
Here, $\theta$ is a shallow projection used only during training. The edge target regularizes latent-derived conditioning; neither the target edge map nor full-resolution encoder features are supplied directly to the PET decoder. Fig.~\ref{fig:sfl} shows the structural pathway.

\subsection{Total Training Objective}
The SFL-Net objective is formulated by combining PET reconstruction, vector-quantization regularization, source factorization, structural consistency, and anti-collapse regularization:
\begin{align}
\mathcal{L}={}&\lambda_{\mathrm{PET}}\mathcal{L}_{\mathrm{PET}}
+\lambda_{\mathrm{VQ}}\mathcal{L}_{\mathrm{VQ}}
+\lambda_{\mathrm{SF}}\mathcal{L}_{\mathrm{SF}} \nonumber\\
&+\lambda_{\mathrm{edge}}\mathcal{L}_{\mathrm{edge}}
+\lambda_{\mathrm{vf}}\mathcal{L}_{\mathrm{vf,tot}}.
\end{align}
Here, $\mathcal{L}_{\mathrm{PET}}$ is the selected PET reconstruction objective, $\mathcal{L}_{\mathrm{VQ}}$ is the vector-quantization regularizer \cite{oord2017vqvae}, $\mathcal{L}_{\mathrm{SF}}$ is the source-factorization objective, $\mathcal{L}_{\mathrm{edge}}$ supervises latent structural conditioning, and $\mathcal{L}_{\mathrm{vf,tot}}$ discourages collapse in the source-specific and cross-source pathways. The weighting coefficients balance the reconstruction and regularization terms, with $\lambda_{\mathrm{align}}=1$.

\section{Experiments}

\subsection{Data and Preprocessing}
We evaluated SFL-Net using ADNI-3 \cite{Weiner2017} and OASIS-3 \cite{LaMontagne2019}. After preprocessing and quality control, 605 subjects were assigned to training and 83 subjects to an internal validation set. The split was subject-level, but subjects from the two cohorts were pooled before random partitioning. Site, scanner, acquisition protocol, demographic, and cohort metadata were not used for training or stratified evaluation. Consequently, the validation set should be interpreted as an internal pooled-cohort evaluation rather than an external cohort-, site-, or scanner-held-out test.

For each subject, FLAIR MRI and tau-PET were registered to the corresponding T1-weighted MRI space. The maximum permitted acquisition-time difference was 1.5 years. This design evaluates a temporally heterogeneous association: prediction discrepancies may combine synthesis error, biological change, registration/resampling effects, and measurement variability. Their contributions cannot be separated here, and no interval-stratified analysis was performed. Volumes were skull stripped using FreeSurfer SynthStrip \cite{hoopes2022synthstrip,kelley2024boosting,hoffmann2025domain}, denoised, cropped, resized, and intensity normalized. Training used on-the-fly random affine transformations, axis flips, and random resized crops.

PET-derived SUVR maps were obtained by normalizing tau-PET uptake to cerebellar cortex uptake. SUVR-domain evaluation uses cerebellar normalization. Regional and staging metrics are interpreted as descriptive summaries rather than evidence of a calibrated MRI-only quantitative pipeline. Braak-relevant regional masks were generated with SynthSeg \cite{BILLOT2023102789} and grouped into Braak I/II, III/IV, and V/VI regions. Five ordered image-derived stages were assigned using the sequential rule \cite{chen2021staging}
{\small
\begin{equation}
\mathrm{Stage}=\begin{cases}
4, & \mathrm{SUVR}_{\mathrm{Braak~V/VI}}>1.873,\\
3, & \mathrm{SUVR}_{\mathrm{Braak~III/IV}}>1.523,\\
2, & \mathrm{SUVR}_{\mathrm{Braak~III/IV}}>1.307,\\
1, & \mathrm{SUVR}_{\mathrm{Braak~I/II}}>1.129,\\
0, & \mathrm{otherwise}.
\end{cases}
\end{equation}}
The validation distribution was Stage~0: 16, Stage~1: 45, Stage~2: 7, Stage~3: 10, and Stage~4: 5. The first satisfied condition in the displayed top-down rule determines the label; positivity in all preceding anatomical regions is not required by this rule. These labels are adapted threshold-based imaging summaries, not neuropathological Braak stages. Transfer of the published cutoffs to this processing pipeline has not been validated. Tracer, partial-volume correction, reference region, parcellation, and regional weighting can affect their applicability.

\subsection{Implemented Comparators, Metrics, and Statistical Analysis}
We compared SFL-Net with VAE, VQ-VAE, U-Net, and SPADE-U-Net variants trained under matched T1+FLAIR input and tau-PET target conditions \cite{Kingma2013AutoEncodingVB,oord2017vqvae,ronneberger2015unet,park2019spade}. All models used the same MKConv backbone where applicable and were capacity matched to approximately five million learnable parameters. Training used an effective batch size of 8 with gradient accumulation and 64 randomly sampled training cases per epoch. Models were trained for 1000 epochs with AdamW and cosine learning-rate decay from $10^{-4}$ to $10^{-5}$. Additional contrast-weighted and gradient-aware variants explored reconstruction-loss shaping \cite{charbonnier1994two,mathieu2015deep,brats2022-transmorph}.

These comparisons should be interpreted narrowly. The experiment did not include a contemporary transformer, GAN, diffusion, or task-specific tau-PET baseline, despite such methods being relevant to the current synthesis literature \cite{dayarathna2024review,li2024pasta,moon2026cyclic}. Nor were matched T1-only and FLAIR-only models trained. Accordingly, the current data do not establish state-of-the-art synthesis performance or the incremental predictive value of T1+FLAIR over either contrast alone.

PET-intensity reconstruction was evaluated using MAE, MSE, PSNR, SSIM, and MS-SSIM \cite{5596999}. PET intensity refers to the pre-SUVR evaluation domain and should not be interpreted as calibrated activity units. SUVR-domain comparisons are restricted to SSIM, MS-SSIM, and regional relative error. Regional relative error was computed within each subject and averaged across staging regions before paired subject-level testing. Regional signed SUVR bias was computed by pooling subject--ROI pairs. Because ROIs from the same subject are correlated, pooled bias is descriptive; conventional pooled limits of agreement are not interpreted as valid repeated-measures Bland--Altman limits \cite{BlandAltman1986}.

Braak-derived staging was summarized by exact accuracy, within-one-stage accuracy, mean absolute stage error (MASE), and quadratic weighted kappa (QWK), both over all subjects and over the Stage~0--2 subset. To expose the effect of stage imbalance, we additionally report a no-image constant Stage~1 reference computed directly from the validation-stage distribution. This is an empirical imbalance reference chosen with knowledge of the evaluation labels, not a prospectively selected clinical comparator. Continuous paired model metrics were compared with two-sided Wilcoxon signed-rank tests, paired binary staging outcomes with exact McNemar tests, and latent-component Shapley contributions with paired Wilcoxon signed-rank tests with Holm correction for pairwise component comparisons. This does not establish control of multiplicity across all utilities. Model-comparison $p$-values are treated as nominal and exploratory because no across-model/metric correction is specified. The reported point estimates do not quantify uncertainty across training runs or alternative splits. Paired subject tests condition on the fitted models and do not measure seed or split variability. Nonsignificance does not establish equivalence.

The validation set was reused to compare model/loss configurations; the primary configuration and displayed comparators should not be treated as prespecified confirmatory choices. There is no independent confirmatory test set.

\subsection{Conditional Latent-Component Attribution}
For latent attribution, the four players were $\mathcal{N}=\{z^s,\allowbreak z^u_{\mathrm{T1}},\allowbreak z^u_{\mathrm{FL}},\allowbreak z^c\}$. The attribution intervention retained subject-specific structural context $h_j$ and replaced absent factors by subject-independent mean latent values. Thus the empty coalition is \emph{retained structure plus mean-replaced latent inputs}, not an information-free or literally structure-only prediction. For subject $j$, let $v_j(S\mid h_j)$ denote the selected utility under coalition $S$. Exact enumeration uses all 16 coalitions:
\begin{equation}
\begin{aligned}
\phi_{j,i}={}&\sum_{S\subseteq\mathcal{N}\setminus\{i\}}\frac{|S|!(3-|S|)!}{4!}\\
&\quad\times\bigl[v_j(S\cup\{i\}\mid h_j)-v_j(S\mid h_j)\bigr].
\end{aligned}
\end{equation}
Error utilities are sign-reversed so that positive contributions indicate improvement \cite{shapley1953value,lundberg2017shap,ren2025comesexplanationshapleyperspective}. A negative value indicates reduced utility on average over coalitions under this intervention, not negative biological information. Mean replacement can create latent combinations outside the learned joint distribution and does not emulate realistic removal of an MRI contrast.

The retained context is derived from shared latent channels. Replacing the main shared player therefore does not remove all shared input-derived information. The shared-player contribution is conditional on the retained structural route and cannot be interpreted as total reliance on shared MRI information. Attribution magnitude also depends on the replacement baseline and utility scaling; it is not a percentage of biological source information.

For the untransformed game, efficiency requires
\begin{equation}
\sum_i\phi_{j,i}=v_j(\mathcal{N}\mid h_j)-v_j(\varnothing\mid h_j).
\end{equation}
This identity provides an arithmetic consistency check for the specified coalition game. Its interpretation concerns the accounting of utility contributions; biological meaning requires a separate argument. Stage-based utilities additionally require acquired-PET labels and constitute retrospective evaluation; their interpretation does not transfer directly to MRI-only deployment.

\begin{table*}[t]
\centering\small
\caption{Principal descriptive reconstruction comparison ($n=83$). PET intensity denotes the pre-SUVR evaluation domain. SUVR results summarize image similarity and regional relative error; they do not establish individual quantitative agreement. SF: source-factorization objective; VCC: contrast-weighted Charbonnier; AG: gradient-edge objective. Variants are fitted configurations, not isolated causal effects.}
\label{tab:principal_recon}
\resizebox{\textwidth}{!}{\begin{tabular}{lccccc|ccc}\toprule
&\multicolumn{5}{c|}{PET intensity}&\multicolumn{3}{c}{SUVR summaries}\\
Model & MAE$\downarrow$ & MSE$\downarrow$ & PSNR$\uparrow$ & SSIM$\uparrow$ & MS-SSIM$\uparrow$ & SSIM$\uparrow$ & MS-SSIM$\uparrow$ & RelErr (\%)$\downarrow$\\\midrule
VAE (MSE) & 0.1243 & 0.0212 & 16.65 & 0.838 & 0.747 & 0.841 & 0.747 & 46.3\\
VQ-VAE (MSE) & 0.1215 & 0.0210 & 16.91 & 0.841 & 0.747 & 0.844 & 0.751 & 45.3\\
U-Net (MSE) & 0.1201 & 0.0209 & 17.07 & 0.854 & 0.812 & 0.855 & 0.812 & 31.2\\
SPADE-U-Net (MSE) & 0.1072 & 0.0178 & 18.31 & 0.861 & 0.850 & 0.866 & 0.857 & 12.8\\
SPADE-U-Net (VCC) & 0.1066 & 0.0175 & 18.36 & 0.863 & 0.854 & 0.868 & 0.861 & 13.4\\
SPADE-U-Net (MSE) + AG & 0.1072 & 0.0178 & 18.29 & 0.863 & 0.851 & 0.868 & 0.857 & 13.7\\
SFL-Net (MSE) & 0.1085 & 0.0178 & 18.22 & 0.861 & 0.842 & 0.865 & 0.848 & 15.7\\
SFL-Net (SF) (MSE) & 0.1043 & 0.0168 & 18.53 & 0.862 & 0.848 & 0.866 & 0.855 & 13.1\\
\bottomrule\end{tabular}}
\end{table*}

\begin{table*}[t]
\centering\small
\caption{Descriptive regional bias and adapted threshold-derived staging ($n=83$; early subset $n=68$). Bias is pooled over correlated subject--ROI pairs; it does not establish individual agreement. Stage~1 is an empirical no-image imbalance reference. Threshold transfer and synthetic-reference processing are unvalidated. Model labels follow Table~\ref{tab:principal_recon}.}
\label{tab:principal_clinical}
\resizebox{\textwidth}{!}{\begin{tabular}{lcc|cccc|cccc}\toprule
&\multicolumn{2}{c|}{Pooled bias}&\multicolumn{4}{c|}{All stages}&\multicolumn{4}{c}{Stages 0--2}\\
Model & SUVR & \% & Exact$\uparrow$ & Within-1$\uparrow$ & MASE$\downarrow$ & QWK$\uparrow$ & Exact$\uparrow$ & Within-1$\uparrow$ & MASE$\downarrow$ & QWK$\uparrow$\\\midrule
VAE (MSE) & -0.288 & -18.4 & 0.193 & 0.735 & 1.313 & 0.000 & 0.235 & 0.897 & 0.868 & 0.000\\
VQ-VAE (MSE) & -0.283 & -18.1 & 0.193 & 0.747 & 1.301 & 0.006 & 0.235 & 0.912 & 0.853 & 0.031\\
U-Net (MSE) & -0.286 & -18.4 & 0.193 & 0.735 & 1.313 & 0.000 & 0.235 & 0.897 & 0.868 & 0.000\\
SPADE-U-Net (MSE) & -0.132 & -5.8 & 0.494 & 0.819 & 0.819 & -0.067 & 0.603 & 0.985 & 0.412 & 0.108\\
SPADE-U-Net (VCC) & -0.106 & -3.5 & 0.530 & 0.831 & 0.735 & -0.051 & 0.647 & 1.000 & 0.353 & 0.135\\
SPADE-U-Net (MSE) + AG & -0.102 & -3.1 & 0.554 & 0.831 & 0.723 & -0.002 & 0.676 & 0.985 & 0.338 & 0.165\\
SFL-Net (MSE) & -0.146 & -6.8 & 0.482 & 0.807 & 0.843 & 0.007 & 0.588 & 0.985 & 0.426 & 0.189\\
SFL-Net (SF) (MSE) & -0.101 & -3.3 & 0.518 & 0.831 & 0.711 & 0.140 & 0.632 & 0.985 & 0.382 & 0.216\\
Constant Stage~1 & -- & -- & 0.542 & 0.819 & 0.699 & 0.000 & 0.662 & 1.000 & 0.338 & 0.000\\
\bottomrule\end{tabular}}
\end{table*}

\section{Results}

\subsection{Descriptive Reconstruction Performance}
Tables~\ref{tab:principal_recon} and \ref{tab:principal_clinical} summarize the internal comparisons; Supplementary Tables~S1a and~S1b retain the full configuration grid for interpretable metrics. SFL-Net (SF + MSE) had the lowest reported PET-intensity MAE (0.1043) and MSE (0.0168), and the highest reported PSNR (18.53 dB). SPADE-U-Net (MSE) had MAE 0.1072 and PSNR 18.31 dB, whereas SPADE-U-Net (VCC) had MAE 0.1066 and PSNR 18.36 dB. These are small differences in reported point estimates, not evidence of practical superiority or equivalence.

For the SF-on/off configurations, MAE was 0.1043 versus 0.1085 and PSNR was 18.53 versus 18.22 dB. Nominal paired tests gave $p\leq0.0151$ for all PET-intensity reconstruction metrics. This comparison characterizes the contribution of the regularized configuration in the evaluated runs. Its interpretation is conditional on the fitted models and the chosen validation partition.

SUVR similarity and regional relative-error point estimates favored some SPADE-U-Net variants: the MSE variant's reported regional relative error was 12.8\% versus 13.1\% for primary SFL-Net. Nominal paired comparisons between primary SFL-Net and SPADE-U-Net (MSE + AG) favored SFL-Net on PET-intensity MSE and PSNR ($p=0.0110$ and $p=0.0171$) and the comparator on SUVR SSIM and MS-SSIM ($p=0.0424$ and $p=0.0086$). These tests are interpreted as nominal comparisons without across-metric correction and do not establish confirmatory superiority.

The qualitative examples (Supplementary Fig.~S1) show smooth synthesized uptake patterns and attenuation of focal high signal. These examples illustrate possible errors rather than their population-level prevalence. Neither smoothing nor the metric differences establish a fidelity--interpretability trade-off caused by removal of feature skips.

\begin{figure*}[t]
\centering
\includegraphics[width=0.92\textwidth]{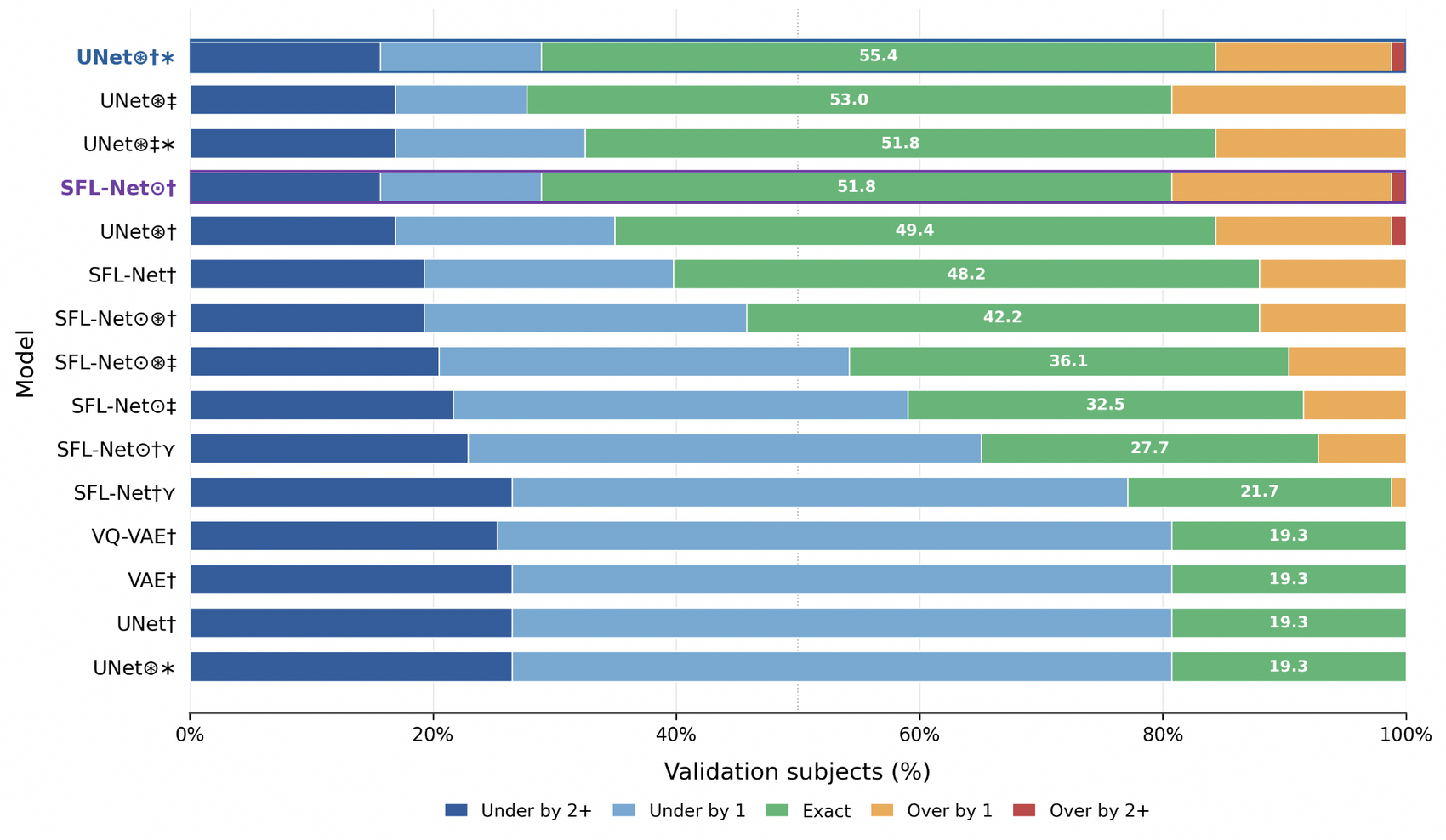}
\caption{Signed Braak-derived stage-error composition across the evaluated synthesis models. Bars show the fraction of validation subjects underestimated by two or more stages, underestimated by one stage, exactly matched, overestimated by one stage, or overestimated by two or more stages. The percentage printed within the green segment is exact-stage agreement. The plot excludes the constant Stage~1 reference; Table~\ref{tab:principal_clinical} provides that essential context. Stages are adapted imaging categories, not neuropathological labels. Symbols identify MSE ($\dagger$), VCC ($\ddagger$), AG ($*$), SPADE ($\circledast$), SF ($\odot$), and ResNet supervision ($\curlyvee$).}
\label{fig:braak_signed_error}
\end{figure*}

\subsection{Regional Bias, Dynamic-Range Compression, and Staging}
Among the evaluated image models, primary SFL-Net had the smallest reported absolute pooled signed regional SUVR bias ($-0.101$ SUVR), while U-Net\spademark\msemark\agmark had the smallest absolute percent bias ($-3.1\%$ versus $-3.3\%$ for SFL-Net). These pooled averages should not be interpreted as repeated-measures limits of agreement because each subject contributes multiple correlated ROIs. The regional summaries and examples suggest magnitude-dependent error, with attenuation at higher uptake and higher imaging stages. The prevalence and subgroup dependence of this pattern were not evaluated. A small global signed bias can therefore coexist with substantial regional disagreement because positive and negative errors cancel.

Braak-derived staging further exposes the effect of imbalance and dynamic-range compression. SFL-Net\sfmark\msemark achieved exact accuracy 0.518, within-one-stage accuracy 0.831, MASE 0.711, and QWK 0.140. A no-image predictor that assigns every validation subject to the majority Stage~1 category obtains exact accuracy $45/83=0.542$, within-one-stage accuracy $68/83=0.819$, MASE $58/83=0.699$, and QWK 0.000. In the Stage~0--2 subset, the same constant reference obtains exact accuracy $45/68=0.662$, within-one-stage accuracy 1.000, and MASE $23/68=0.338$, compared with 0.632, 0.985, and 0.382 for SFL-Net. Thus, exact accuracy, within-one-stage accuracy, and MASE are strongly influenced by the validation-stage distribution and do not provide evidence of useful staging performance in isolation. SFL-Net's positive QWK indicates some ordinal association beyond a constant prediction, but its magnitude remains low. Within-one-stage agreement differs from the constant reference by approximately one subject, whereas exact agreement and MASE are worse. We therefore treat staging as a failure-oriented threshold-sensitivity analysis, not a demonstrated downstream clinical capability or a formal calibration assessment.

\subsection{Conditional Attribution: Qualitative Findings and Limits}
The exploratory attribution visualizations (Supplementary Figs.~S2 and~S3) suggest greater average contribution from the joint-input branch across reconstruction and stage-derived utilities, with substantial subject variation. Negative contributions in some subjects indicate that this branch does not uniformly improve the selected utility. The T1-only-input pathway contributes little or negatively on average for some stage utilities. This does not imply that T1 is dispensable: it also enters shared and joint-input routes.

These patterns characterize conditional branch dependence. We do not interpret contribution magnitudes as fractions of source information or compare them across utility scales. Associations between contribution heterogeneity and clinical, anatomical, or acquisition characteristics were not evaluated. The resulting distributions provide a subject-level view of decoder reliance on the prescribed branches. They answer a model-inspection question under latent replacement, rather than assigning biological importance to an MRI contrast.

\section{Discussion}
SFL-Net combines predefined latent intervention units with a two-input synthesis model. Its central contribution connects a structured bottleneck, latent-derived anatomical conditioning, and coalition-based model inspection within a common framework. The internal comparison demonstrates favorable PET-intensity errors for the primary configuration, while regional and staging summaries provide a complementary view of its prediction behavior. The resulting analysis connects architectural design with subject-level variation in decoder reliance.

\subsection{Positioning and Meaning of the Latent Routes}
Common and modality-specific representations have clear precedents. Hi-Net combines modality-specific and common multimodal representations, and feature-disentanglement approaches separate modality-invariant and modality-specific spaces \cite{zhou2020hinet,fei2021disentanglement}. SFL-Net combines prescribed vector-quantized pathways, pairwise dependence regularization, a joint-input encoder, latent-derived structural conditioning, and coalition attribution. These ingredients do not constitute a new information decomposition. Other architectures, including models with skips, can also be interrogated if their intervention units are defined.

Low pairwise dependence does not prove semantic uniqueness, and a variance floor does not exclude irrelevant high-variance features. A strong joint-branch contribution describes decoder reliance under one replacement scheme, not formal synergy. Retained structural information can reduce the apparent contribution assigned to the main shared route. Interpretation follows the specified intervention: contributions quantify marginal utility relative to the retained structural context and replacement baseline. They are not an exhaustive decomposition of MRI information.

\subsection{Scope of the Comparison}
The SF-on/off pair provides a narrow comparison of fitted configurations. The study lacks matched continuous-bottleneck, restored-skip, structural-route-removal, and single-contrast controls. Thus, neither the necessity of vector quantization nor the predictive benefit of FLAIR can be isolated. A small T1-only-input branch attribution cannot justify omitting T1, and a strong joint-input attribution cannot justify acquiring FLAIR. Missing-modality robustness is untested.

Modern synthesis includes transformer, adversarial, and diffusion approaches \cite{dayarathna2024review,li2024pasta,moon2026cyclic}. Their published numbers are not interchangeable with results under the present split and processing protocol. No such comparator was evaluated here; parameter matching within the implemented model families does not establish contemporary state-of-the-art performance.

The observed smoothing may reflect several aspects of deterministic prediction, training objectives, preprocessing, or architecture. Without the relevant controls, it cannot be attributed to bypass restriction or interpreted as the necessary cost of component attribution. The present study establishes neither a causal fidelity--interpretability trade-off nor a benefit from that trade-off.

\subsection{Threshold Sensitivity and Molecular Interpretation}
The constant Stage~1 reference matches or exceeds primary SFL-Net on exact agreement and mean absolute stage error. Positive but low QWK does not overcome this weak absolute performance. Small pooled signed bias is also insufficient: regional positive and negative errors can cancel, and multiple ROIs per subject are correlated. The results do not establish individual quantitative agreement.

SUVR depends on both regional uptake and reference-region processing, while threshold calibration depends on the tracer and processing convention. A reference computed from synthetic PET is subject to prediction error; using a measured PET reference would introduce target-derived information at evaluation. Regional and staging summaries therefore require careful distinction from an MRI-only quantitative measurement. In all cases, a threshold summary derived from synthetic PET remains a prediction from MRI rather than an independent molecular observation. This study does not demonstrate added diagnostic information beyond MRI or a patient-management benefit.

\subsection{Study Scope and Interpretation}
All comparisons use one 83-subject internal pooled-cohort partition, reused across model and loss configurations. There is no untouched confirmatory test. Cohort, scanner, site, protocol, and demographic effects were not stratified, and their absence as explicit model inputs does not remove potential confounding. Stages~2 and~4 contain only 7 and 5 subjects. Neither repeated training nor alternative splits quantify model-ranking stability. Subject-paired tests do not address those uncertainties.

MRI and tau-PET acquisitions could differ by up to 1.5 years. Progression and selection effects may contribute systematic mismatch rather than random noise alone. Without interval-stratified evaluation, the magnitude of this effect is unknown and specific failures cannot be assigned to temporal mismatch.

SUVR MAE and MSE are excluded from the principal tables because their values cannot be reconciled under a common error definition; SUVR PSNR and high-uptake overlap measures also lack sufficiently specified evaluation conventions. The supplementary performance overview is retained for documentation and is not used to support claims about these metrics. The principal tables therefore emphasize the retained reconstruction, regional, and staging summaries. Attribution displays are interpreted within the stated replacement intervention and utility conventions.

Finally, MRI depicts indirect structural correlates rather than tau tracer binding. Learned associations need not transfer to early molecular disease, alternative causes of atrophy, or different populations. These are plausible boundaries, not measured subgroup results. Outputs remain synthetic research estimates, without demonstrated diagnostic substitutability.

\section{Conclusion}
SFL-Net integrates vector-quantized latent pathways, dependence regularization, and structural conditioning in a framework for multi-contrast MRI-to-tau-PET synthesis. Its predefined shared, single-source, and joint-input routes make the trained decoder amenable to systematic component interrogation. In the internal comparison, the primary configuration achieved the lowest PET-intensity MAE and MSE and the highest PSNR among the implemented models, while the regional and staging summaries revealed performance differences that global reconstruction metrics alone would not capture. Coalition attribution indicated the largest average contribution from the joint-input pathway, together with substantial variation across subjects. Taken together, these findings illustrate the value of coupling structured latent representations with an explicit intervention-based analysis. The framework offers a concrete basis for studying how multimodal predictors distribute computation across latent routes and how that dependence relates to reconstruction behavior.

\FloatBarrier
\section*{Acknowledgments}
\addcontentsline{toc}{section}{Acknowledgments}
This work was supported in part by the U.S. Department of Energy, Office of Science, Office of Advanced Scientific Computing Research, DOE Computational Science Graduate Fellowship under Award No. DE-SC0024386 to Juampablo E. Heras Rivera.

The authors thank Ilka M. Lorenzo of Northwestern University for correspondence concerning the explanation of tau protein. The authors also thank Hershdeep Singh Chopra, affiliated with Northwestern University at the time of those discussions and now with the London School of Economics, for discussions of partial information decomposition that inspired the initial idea for this study.

\FloatBarrier
\balance
\bibliographystyle{IEEEtran}
\bibliography{references}
\clearpage
\begin{table*}[p]
\centering
\section*{Supplementary material}
\small The complete retained configuration grid and extended figure documentation complement the main article. Figures~S2--S4 contain incompletely specified quantities; their captions state the resulting interpretation limits. No additional analyses are introduced.
\par\medskip
\textbf{Table S1a. Complete configuration grid: reconstruction}\par\smallskip
\scriptsize PET intensity denotes the pre-SUVR domain. SUVR summaries report image similarity and regional relative error. Rankings are descriptive because validation was reused across configurations.
\par\smallskip
\par\noindent\resizebox{\textwidth}{!}{\begin{tabular}{lccccc|ccc}\toprule
&\multicolumn{5}{c|}{PET intensity}&\multicolumn{3}{c}{SUVR}\\
Model & MAE & MSE & PSNR & SSIM & MS-SSIM & SSIM & MS-SSIM & RelErr (\%)\\\midrule
VAE\msemark & 0.1243 & 0.0212 & 16.65 & 0.838 & 0.747 & 0.841 & 0.747 & 46.3\\
VQ-VAE\msemark & 0.1215 & 0.0210 & 16.91 & 0.841 & 0.747 & 0.844 & 0.751 & 45.3\\
U-Net\msemark & 0.1201 & 0.0209 & 17.07 & 0.854 & 0.812 & 0.855 & 0.812 & 31.2\\
U-Net\spademark\msemark & 0.1072 & 0.0178 & 18.31 & 0.861 & 0.850 & 0.866 & 0.857 & 12.8\\
U-Net\spademark\agmark & 0.4219 & 0.2472 & 7.52 & 0.688 & 0.002 & 0.729 & 0.020 & 73.0\\
U-Net\spademark\vccmark & 0.1066 & 0.0175 & 18.36 & 0.863 & 0.854 & 0.868 & 0.861 & 13.4\\
U-Net\spademark\msemark\agmark & 0.1072 & 0.0178 & 18.29 & 0.863 & 0.851 & 0.868 & 0.857 & 13.7\\
U-Net\spademark\vccmark\agmark & 0.1063 & 0.0175 & 18.37 & 0.862 & 0.850 & 0.867 & 0.857 & 13.2\\
SFL-Net\msemark & 0.1085 & 0.0178 & 18.22 & 0.861 & 0.842 & 0.865 & 0.848 & 15.7\\
SFL-Net\msemark\rmark & 0.1135 & 0.0188 & 17.65 & 0.852 & 0.811 & 0.855 & 0.810 & 44.6\\
SFL-Net\sfmark\msemark\rmark & 0.1093 & 0.0181 & 18.13 & 0.858 & 0.837 & 0.861 & 0.840 & 26.6\\
SFL-Net\sfmark\msemark & 0.1043 & 0.0168 & 18.53 & 0.862 & 0.848 & 0.866 & 0.855 & 13.1\\
SFL-Net\sfmark\vccmark & 0.1088 & 0.0182 & 18.06 & 0.855 & 0.835 & 0.859 & 0.839 & 25.6\\
SFL-Net\sfmark\spademark\msemark & 0.1093 & 0.0181 & 18.12 & 0.856 & 0.831 & 0.860 & 0.837 & 16.6\\
SFL-Net\sfmark\spademark\vccmark & 0.1084 & 0.0180 & 18.12 & 0.857 & 0.839 & 0.860 & 0.843 & 23.1\\
\bottomrule\end{tabular}}

$\dagger$: MSE; $\ddagger$: contrast-weighted Charbonnier (VCC); $*$: gradient-edge (AG); $\circledast$: SPADE; $\odot$: source-factorization objective (SF); $\curlyvee$: ResNet-18 supervision. Configurations with different objectives or supervision are not isolated architectural controls.\par\medskip
\textbf{Table S1b. Complete configuration grid: regional bias and staging}\par\smallskip
\scriptsize Pooled bias is descriptive. Stages are adapted imaging categories; threshold transfer is unvalidated. Early-stage metrics use 68 subjects in categories 0--2. MASE denotes mean absolute stage error. The constant Stage~1 reference is derived from the validation distribution.
\par\smallskip
\par\noindent\resizebox{\textwidth}{!}{\begin{tabular}{lcc|cccc|cccc}\toprule
&\multicolumn{2}{c|}{Pooled bias}&\multicolumn{4}{c|}{All stages ($n=83$)}&\multicolumn{4}{c}{Stages 0--2 ($n=68$)}\\
Model & SUVR & \% & Exact & Within-1 & MASE & QWK & Exact & Within-1 & MASE & QWK\\\midrule
VAE\msemark & -0.288 & -18.4 & 0.193 & 0.735 & 1.313 & 0.000 & 0.235 & 0.897 & 0.868 & 0.000\\
VQ-VAE\msemark & -0.283 & -18.1 & 0.193 & 0.747 & 1.301 & 0.006 & 0.235 & 0.912 & 0.853 & 0.031\\
U-Net\msemark & -0.286 & -18.4 & 0.193 & 0.735 & 1.313 & 0.000 & 0.235 & 0.897 & 0.868 & 0.000\\
U-Net\spademark\msemark & -0.132 & -5.8 & 0.494 & 0.819 & 0.819 & -0.067 & 0.603 & 0.985 & 0.412 & 0.108\\
U-Net\spademark\agmark & -0.886 & -68.2 & 0.193 & 0.735 & 1.313 & 0.000 & 0.235 & 0.897 & 0.868 & 0.000\\
U-Net\spademark\vccmark & -0.106 & -3.5 & 0.530 & 0.831 & 0.735 & -0.051 & 0.647 & 1.000 & 0.353 & 0.135\\
U-Net\spademark\msemark\agmark & -0.102 & -3.1 & 0.554 & 0.831 & 0.723 & -0.002 & 0.676 & 0.985 & 0.338 & 0.165\\
U-Net\spademark\vccmark\agmark & -0.136 & -5.9 & 0.518 & 0.831 & 0.771 & -0.050 & 0.632 & 1.000 & 0.368 & 0.134\\
SFL-Net\msemark & -0.146 & -6.8 & 0.482 & 0.807 & 0.843 & 0.007 & 0.588 & 0.985 & 0.426 & 0.189\\
SFL-Net\msemark\rmark & -0.236 & -14.2 & 0.217 & 0.735 & 1.289 & -0.019 & 0.265 & 0.897 & 0.838 & -0.013\\
SFL-Net\sfmark\msemark\rmark & -0.196 & -11.0 & 0.277 & 0.771 & 1.133 & 0.005 & 0.338 & 0.941 & 0.721 & -0.003\\
SFL-Net\sfmark\msemark & -0.101 & -3.3 & 0.518 & 0.831 & 0.711 & 0.140 & 0.632 & 0.985 & 0.382 & 0.216\\
SFL-Net\sfmark\vccmark & -0.192 & -10.6 & 0.325 & 0.783 & 1.060 & 0.011 & 0.397 & 0.956 & 0.647 & 0.032\\
SFL-Net\sfmark\spademark\msemark & -0.161 & -8.0 & 0.422 & 0.807 & 0.928 & -0.042 & 0.515 & 0.985 & 0.500 & 0.098\\
SFL-Net\sfmark\spademark\vccmark & -0.180 & -9.4 & 0.361 & 0.795 & 1.024 & -0.040 & 0.441 & 0.971 & 0.588 & 0.069\\
Constant Stage~1 & -- & -- & 0.542 & 0.819 & 0.699 & 0.000 & 0.662 & 1.000 & 0.338 & 0.000\\
\bottomrule\end{tabular}}

$\dagger$: MSE; $\ddagger$: contrast-weighted Charbonnier (VCC); $*$: gradient-edge (AG); $\circledast$: SPADE; $\odot$: source-factorization objective (SF); $\curlyvee$: ResNet-18 supervision. Configurations with different objectives or supervision are not isolated architectural controls.\par\smallskip
\scriptsize Cohort-stratified evaluation, repeated training, and alternative splits were not performed. These point estimates describe fitted-model comparisons on the internal validation partition.
\end{table*}
\clearpage
\setcounter{figure}{0}
\renewcommand{\thefigure}{S\arabic{figure}}
\renewcommand{\theHfigure}{supp.\arabic{figure}}
\begin{figure*}[p]
\centering
\includegraphics[width=0.96\textwidth,height=0.77\textheight,keepaspectratio]{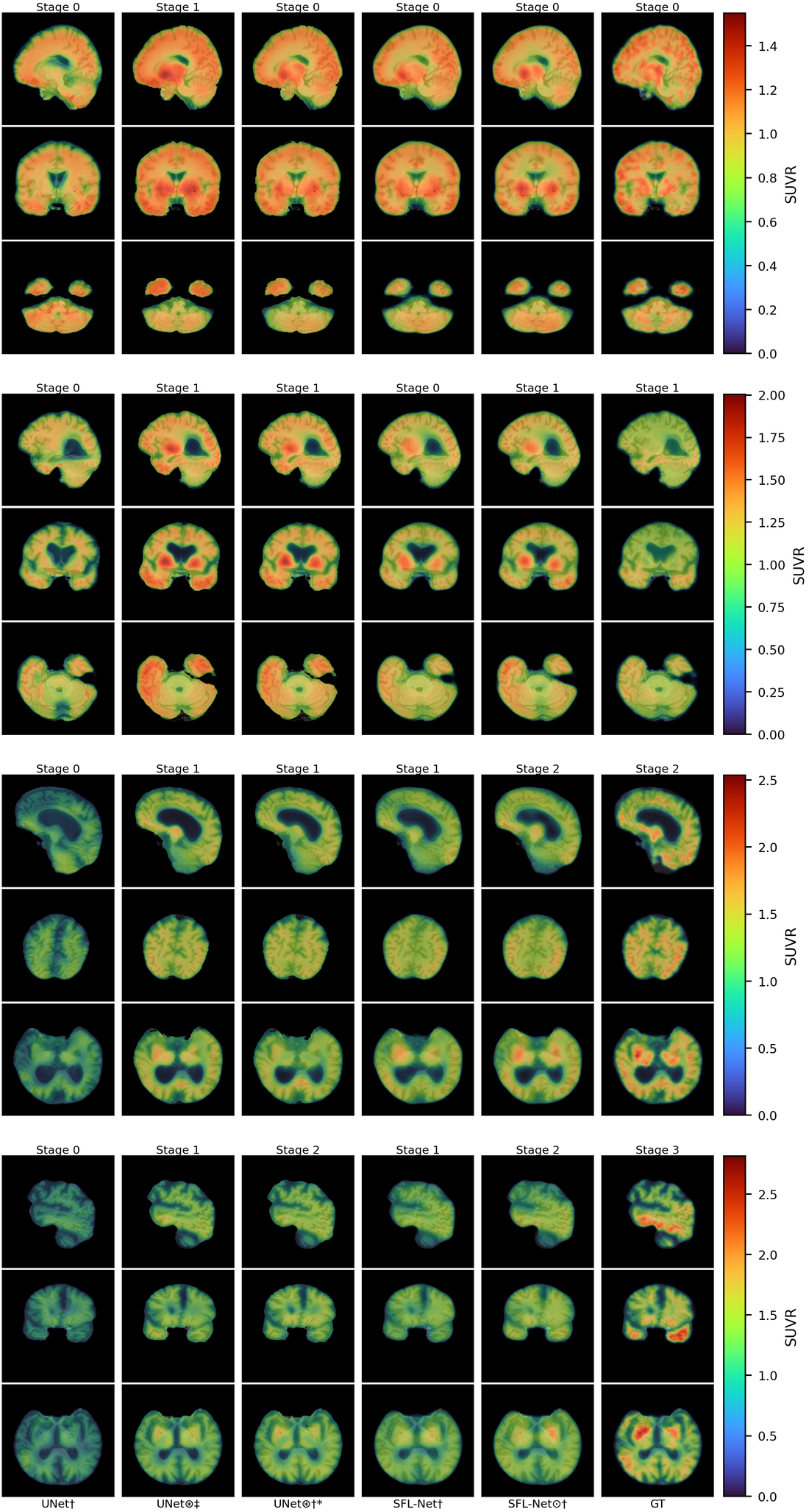}
\caption{Qualitative SUVR reconstructions. Sagittal, coronal, and axial views. Columns: U-Net (MSE), SPADE-U-Net (VCC), SPADE-U-Net (MSE + AG), SFL-Net (MSE), SFL-Net (SF + MSE), and acquired PET (GT). Color scales differ between case blocks. These cases illustrate smoothing and focal attenuation without establishing population prevalence. Stage labels are adapted imaging categories. }
\end{figure*}
\clearpage

\begin{figure*}[p]
\centering
\includegraphics[width=0.96\textwidth,height=0.34\textheight,keepaspectratio]{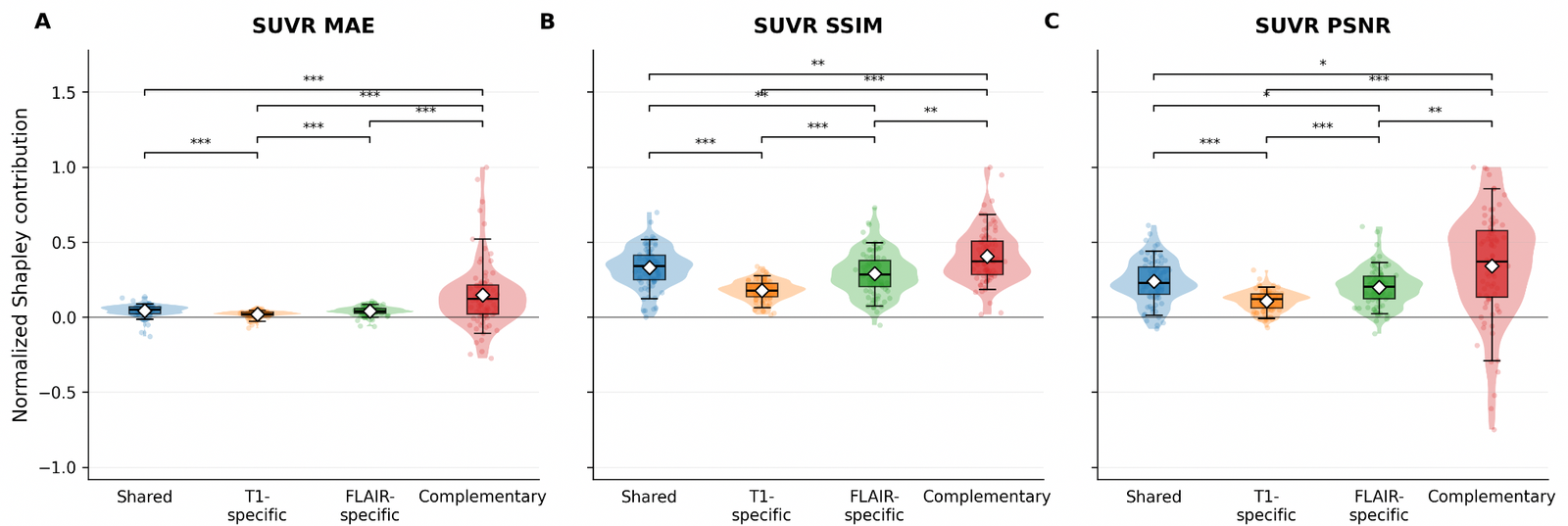}
\caption{Exploratory reconstruction-utility attribution. Subject-level latent-component visualizations ($n=83$). Points, distribution shapes, box summaries, and mean diamonds depict heterogeneous contributions under mean-replacement interventions with retained structural context. ``Shared,'' ``T1-specific,'' ``FLAIR-specific,'' and ``Complementary'' identify the shared-designated, T1-only-input, FLAIR-only-input, and joint-input pathways. Negative contributions indicate reduced utility under the intervention; they do not indicate negative biological information. \textbf{Interpretation.} The plotted normalization has not been specified sufficiently to interpret the vertical axis quantitatively. Values should not be read as percentages of source information or compared across panels. The SUVR MAE and PSNR panels should be read alongside the metric-definition qualifications in the main text. The displayed significance annotations are retained from the original visualization; their correction family and scale are insufficiently specified for confirmatory inference. The figure illustrates subject heterogeneity in conditional decoder reliance; contribution magnitudes should be interpreted within each utility and replacement convention.}
\end{figure*}
\clearpage

\begin{figure*}[p]
\centering
\includegraphics[width=0.96\textwidth,height=0.60\textheight,keepaspectratio]{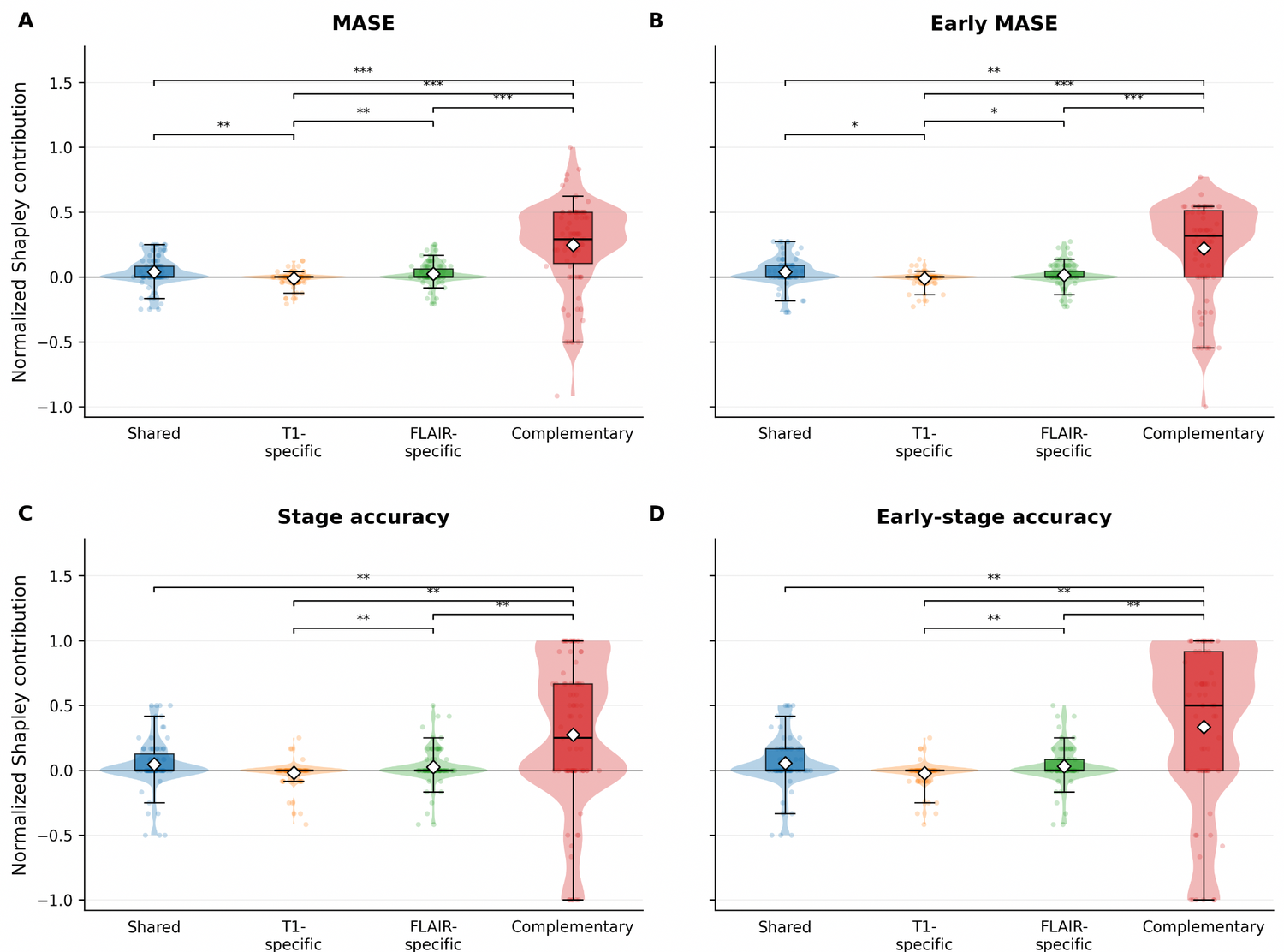}
\caption{Exploratory stage-utility attribution. MASE and stage-accuracy utilities use 83 subjects; early-stage panels use the 68 subjects with acquired-PET categories 0--2. Error utilities are sign-adjusted so that positive contributions indicate improvement. Component labels follow Fig.~S2. \textbf{Interpretation.} These retrospective utilities require acquired-PET labels. The normalization and intervention conventions impose the same quantitative limitations as Fig.~S2; the significance symbols are not used as confirmatory evidence. The displays illustrate heterogeneous decoder reliance rather than clinically useful staging or source-level information fractions.}
\end{figure*}
\clearpage

\begin{figure*}[p]
\centering
\includegraphics[width=0.96\textwidth,height=0.68\textheight,keepaspectratio]{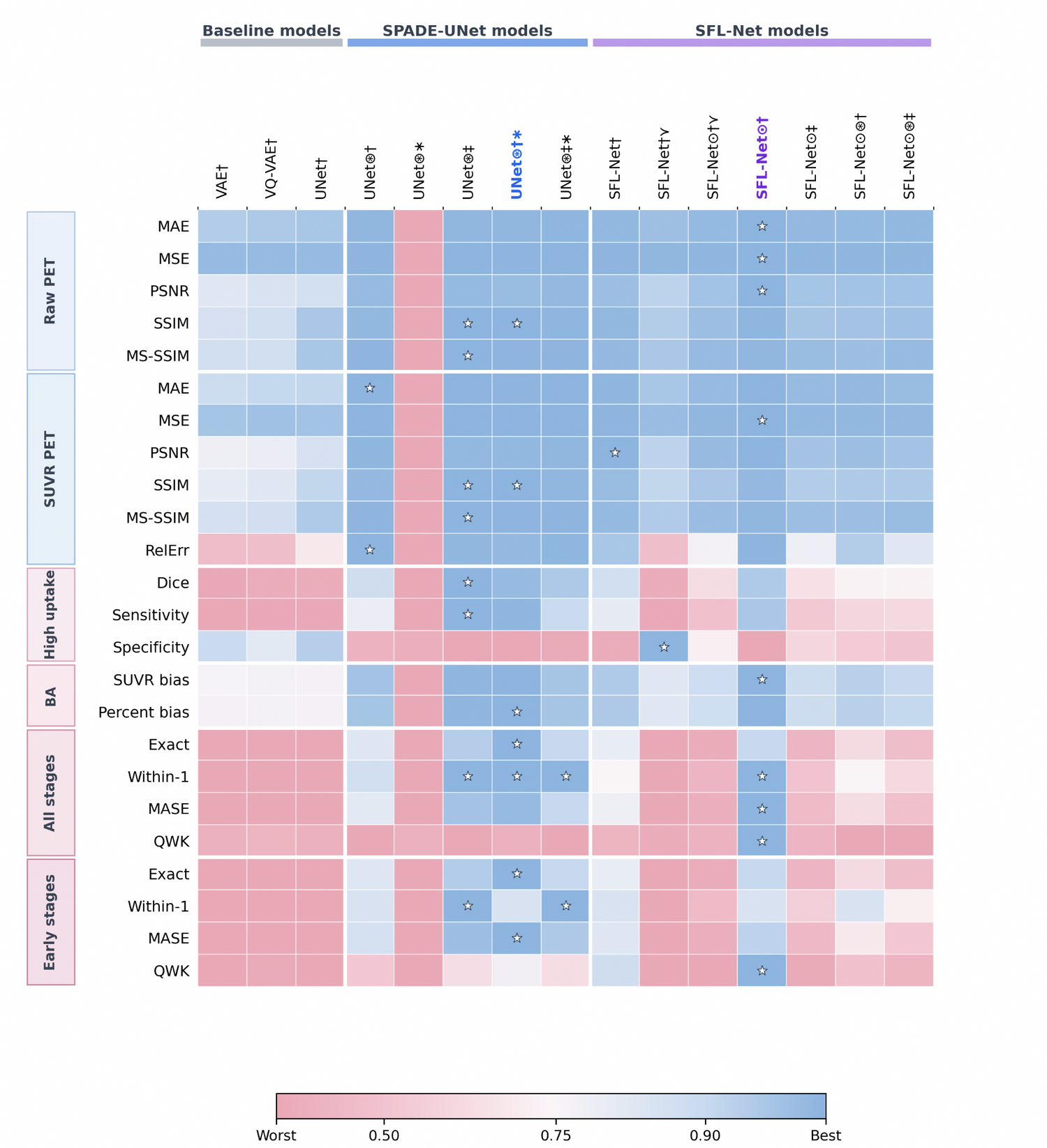}
\caption{Descriptive metric overview. Original overview of the evaluated image-model configurations. Color orientation is worse (pink) to better (blue) within a metric; colors are not a common effect-size scale. Stars mark the original image-model ranking, not statistical significance. ``Raw PET'' denotes the pre-SUVR PET-intensity domain; ``BA'' identifies pooled descriptive bias rather than valid repeated-measures limits of agreement. \textbf{Interpretation.} SUVR MAE/MSE/PSNR and high-uptake metrics are excluded from the principal numerical comparison for the reasons given in the main text. The normalization transform is unspecified, and the constant Stage~1 reference is absent. In particular, the image-model MASE ranking does not outperform the constant reference. Refer to Tables~S1a and~S1b and the main-text limitations for the retained numerical results. This visualization is included for documentation of the full evaluation display, with the principal tables providing the numerical basis for model comparisons.}
\end{figure*}
\clearpage

\end{document}